\documentclass[runningheads]{llncs}
\usepackage[T1]{fontenc}

\usepackage{graphicx}
\usepackage{tabularx}
\usepackage{booktabs}
\usepackage{multirow}
\usepackage{booktabs}
\usepackage{siunitx}
\usepackage{amsmath,amssymb,amsfonts}
\usepackage[table,xcdraw]{xcolor}
\usepackage{hyperref}
\hypersetup{
    colorlinks=true,   
    urlcolor=blue,
}
\usepackage{color, colortbl}
\usepackage{footmisc}
\usepackage{subfig}
\usepackage{float}

\begin{document}

\title{Anatomical Positional Embeddings}

\author{
    Mikhail Goncharov \inst{1,2}$^{\text{,}*\text{,}\dag}$ \and
    Valentin Samokhin \inst{2,3}$^{\text{,}\dag}$ \and
    Eugenia Soboleva \inst{1} \and
    Roman Sokolov \inst{4} \and
    Boris Shirokikh \inst{1,2} \and
    Mikhail Belyaev \inst{5} \and
    Anvar Kurmukov \inst{5}$^{\text{,}\ddag}$ \and
    Ivan Oseledets \inst{1,2}$^{\text{,}\ddag}$
}

\authorrunning{M. Goncharov, V. Samokhin et al.}

\institute{
    Skolkovo Institute of Science and Technology, Moscow, Russia
    \and
    Artificial Intelligence Research Institute (AIRI), Moscow, Russia
    \and
    Institute for Information Transmission Problems, Moscow, Russia
    \and
    Moscow Institute of Physics and Technology, Moscow, Russia
    \and
    Independent researcher
    \\
    $^*$ Correspondence \email{Mikhail.Goncharov2@skoltech.ru}
}

\maketitle

\def\thefootnote{$\dag$}\footnotetext{Equal contribution}
\def\thefootnote{$\ddag$}\footnotetext{Equal supervision}

\begin{abstract}
We propose a self-supervised model producing 3D anatomical positional embeddings (APE) of individual medical image voxels. APE encodes voxels' anatomical closeness, i.e., voxels of the same organ or nearby organs always have closer positional embeddings than the voxels of more distant body parts. In contrast to the existing models of anatomical positional embeddings, our method is able to efficiently produce a map of voxel-wise embeddings for a whole volumetric input image, which makes it an optimal choice for different downstream applications. We train our APE model on 8400 publicly available CT images of abdomen and chest regions. We demonstrate its superior performance compared with the existing models on anatomical landmark retrieval and weakly-supervised few-shot localization of 13 abdominal organs. As a practical application, we show how to cheaply train APE to crop raw CT images to different anatomical regions of interest with 0.99 recall, while reducing the image volume by $10$-$100$ times. The code and the pre-trained APE model are available at \url{https://github.com/mishgon/ape}.

\end{abstract}
\section{Introduction}

Human body has a 3D geometry. Different body parts have more or less the same relative positions in most patients. This inspires so called body part regression models, which aim to learn 3-dimensional positional embeddings encoding anatomical location of different parts of a medical volumetric image. The main idea of learning such models is to predict relative positions of image patches/slices/voxels in 3D physical space based on their appearance.
Networks pretrained in this manner can later be used as few-shot learners for region of interest localization, or applied to image registration, retrieval or lesion tracking. 

Unsupervised body part regressor (BPR) \cite{bpr} and Deep-index \cite{deepindex} proposed algorithms for body part regression in CT images, outputing single number for each 2D axial slice of a volumetric image. BPR uses a two-component loss: one component for the distance between two axial levels, and another for the sign. In turn, Deep-index simplifies loss to a signed distance between two axial levels, demonstrating similar localization performance. While these methods are fast, they provide coarse annotation, only allowing for rough localization, e.g., "lungs are located between axial slice X and Y", and are not accurate for small organs.

Relative Position Regression (RPR) \cite{rpr} could be seen as generalization of \cite{bpr} and \cite{deepindex} from 1D to 3D. Authors train two ResNet-like models ("coarse" and "fine") to predict 3-dimensional positional embeddings of patches, such that offsets between the embeddings coincide with the offsets between the centers of patches in mms. Main drawback of their method comes from a patch-to-vector architecture, making it infeasible to construct a voxel-wise embeddings map.

Self-supervised Anatomical eMbedding (SAM) \cite{sam} is a contrastive method for self-supervised learning of voxel-level representations, capturing both global and local semantics. It outputs two feature maps of high and low resolution, containing multi-scale high-dimensional voxel-level embeddings, which authors adopt for anatomical landmarks retrieval task. High dimensionality of the produced maps is a downside for the image retrieval task due to high storage memory footprint: one SAM map takes $\times 256$ storage space compared to the original image.


\begin{table}[h]
\caption{Main differences between the existing models of anatomical embeddings.}
\centering
\begin{tabularx}{\textwidth}{XXXX}
    \toprule
    \textbf{Paper Citation} & \textbf{Resolution} & \textbf{Embedding size} & \textbf{Pipeline}  \\
    \midrule
    BPR~\cite{bpr} & Axial slices & 1 & single-scale, 1-stage\\
    Deep-index~\cite{deepindex} & Axial slices & 1 & single-scale, 2-stage \\
    RPR~\cite{rpr} & 3D patches &  3 & multi-scale, 2-stage \\
    SAM~\cite{sam} & Voxels & $128 + 128$ & multi-scale, 1-stage \\
    APE (ours) & Voxels & 3 & single-scale, 1-stage \\
    \bottomrule
\end{tabularx}
\label{tab:related_works}
\end{table}

In our work, we propose a novel model producing voxel-level 3-dimensional anatomical positional embeddings. In contrast to RPR~\cite{rpr}, our model is capable of producing positional embeddings for all individual voxels due to its UNet-like architecture. Instead of using two different coarse and fine models as in~\cite{rpr}, we ensure that our positional embeddings capture both global and local anatomical semantics due to our novel training strategy inspired by contrastive methods~\cite{sam}. Thus, our approach bridges the gap between patch-level multi-stage methods~\cite{rpr}, and contrastive methods producing high-dimensional voxel-wise embeddings in a straightforward end-to-end manner~\cite{sam}, see Table~\ref{tab:related_works}.

Our key contributions are two-fold. \textbf{First}, we propose a novel model producing a high-resolution map of voxel-level 3-dimensional anatomical positional embeddings (APE), with naturally emerged properties: continuity with respect to voxels' physical coordinates; orthogonality of embeddings' dimensions; equivariance with respect to input image crops, and changes of its spacing. \textbf{Second}, APE demonstrates a superior performance compared to the existing SotA models on weakly supervised few-shot organ localization, achieving a total localization recall, while reducing image volume $\times10$-$100$ times.

    

\section{Method}
\label{sec:method}

\subsection{APE model}

\begin{figure}[t]
\begin{center}
    \includegraphics[width=1\linewidth]{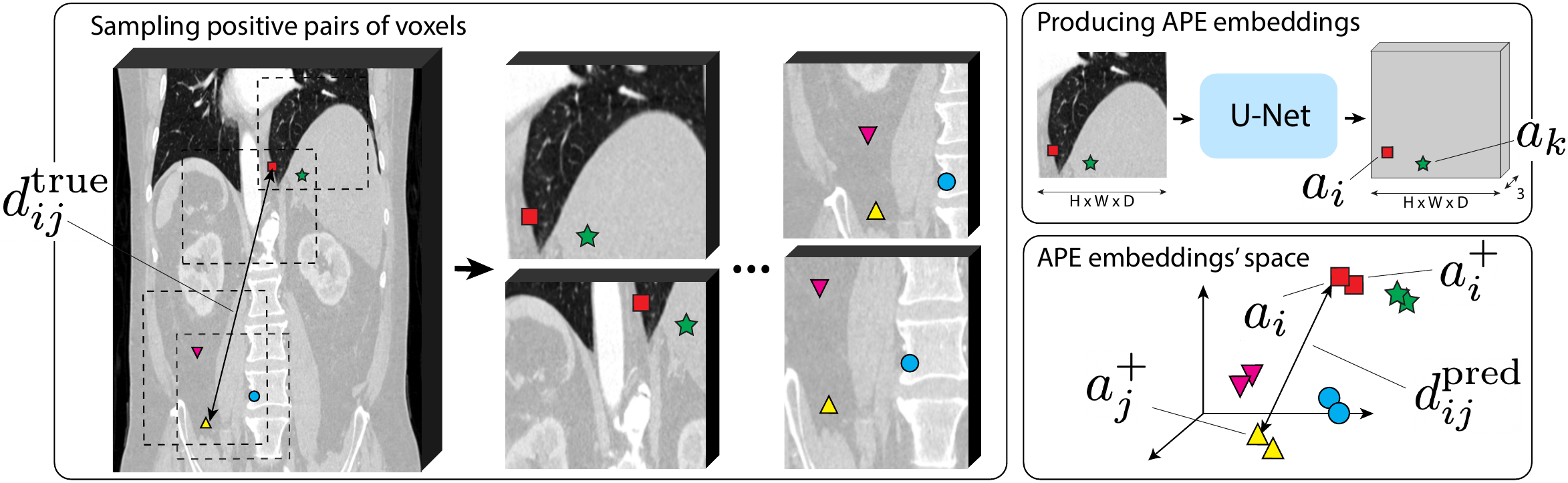}
    \caption{Illustration of APE training procedure. Left: sampling $n$ pairs of overlapping augmented patches and $N$ positive pairs of voxels from the overlapping regions, denoted by the markers of the same color and shape. Top-right: predicting the APE map for each patch and extracting the APE embedding for each sampled voxel. Bottom-right: the relative positions of the positive pairs of voxels in the APE embeddings' space. We train APE such that distances $d_{ij}^{\text{pred}}$ between the APE embeddings are similar to the distances $d_{ij}^{\text{true}}$ between the voxels' normalized absolute positions in the 3D physical space.}
    \label{fig:ape}
\end{center}
\end{figure}

The core idea of APE is to learn how radiological features are located w.r.t. each other in the 3D physical space in an ``average patient''. We implement this idea by sampling different voxels from the same volumetric image and training APE to embed them in a 3D positional embeddings' space based on their local appearances. The main training objective is to make the distances between predicted positional embeddings highly correlated with the physical distances between the corresponding voxels. Although physical distances are defined only between voxels from the same volume, APE embeddings of similar anatomical landmarks in different patients occurs well-aligned with each other. Therefore, it is natural to call them anatomical positional embeddings.

Below, we detail the APE architecture and its training procedure, illustrated in Figure~\ref{fig:ape}.

\paragraph{Architecture} In contrast to RPR~\cite{rpr}, APE model efficiently produces embeddings for all individual voxels of an input image. It is achieved via UNet-like~\cite{nnunet} architecture: it takes a volumetric image patch of size $(H, W, D)$ as input and predicts a tensor of size $(3, H, W, D)$, which we refer to as APE map. This map contains $H \times W \times D$ individual voxels' embeddings of size $3$, called APE embeddings. To produce the APE map for the whole CT image, we use patch-wise prediction. Regarding the network architecture, two details are important: 1) to make the inference fast and memory-efficient we use convolution with stride $4 \times 4 \times 4$ at the first layer and trilinearly upsample the final APE map by a factor of $4$ at the last layer; 2) before the final upsampling layer we use the batch normalization layer without affine rescaling in order to ensure the zero mean and the one standard deviation of APE embeddings at the architectural level.

\paragraph{Naïve training procedure} In its basic form, one APE training iteration is performed as follows. We sample a batch of $n$ volumetric patches from \textit{the same} image and sample $k$ voxels from each patch, resulting in $N = n \cdot k$ voxels in total. Then, we predict their APE embeddings $\{a_i\}_{i = 1}^N \subset \mathbb{R}^3$, and compute the pairwise euclidian distances $d^{\text{pred}}_{ij} = \|a_i - a_j\|_2$. We also compute the voxels' absolute coordinates $\{p_i\}_{i = 1}^N$ w.r.t. the raw image and normalize them to zero mean and unit standard deviation $\{\hat{p}_i\}_{i = 1}^N$. This is done to align them with the predicted embeddings $\{a_i\}_{i = 1}^N$, which are already normalized via batch normalization. We obtain the target pairwise distances between the voxels as $d^{\text{true}}_{ij} = \|\hat{p}_i - \hat{p}_j\|_2$. Then, the training objective is given by:

\begin{equation}
    L_{\text{dist}} = \frac{1}{N^2} \sum_{i, j = 1}^N (d^{\text{pred}}_{ij} - d^{\text{true}}_{ij})^2.
\end{equation}

\noindent
We refer to this most basic APE version as \textit{APE-naïve}. Note that if voxels $i$ and $j$ belong to the different independent patches, the distances $\{d^{\text{true}}_{ij}\}_{i, j = 1}^N$ are often relatively large, so we call them \textit{global}. In opposite, if the voxels $i$ and $j$ belong to the same patch, distances $\{d^{\text{true}}_{ij}\}_{i, j = 1}^N$ are relatively small, and we call them \textit{local}.

Since the network takes a whole patch as input, it can trivially predict the local distances between its voxels without relying on their appearence. Therefore, we believe that \textit{APE-naïve} embeddings correspond to the anatomical positions only on a global scale while being not so well aligned with the anatomical landmarks at local scale, which we empirically show in Section~\ref{sec:results}.

\paragraph{Improved training procedure} In order to achieve the expected local properties of the APE embeddings, such as better alignment with the anatomical landmarks at local scale and equivariance w.r.t. different image crops, changes of its voxel spacing, and color distortions, we propose the following improved training procedure. Inspired by contrastive learning of voxel-level representations~\cite{sam,vox2vec}, we sample $n$ pairs of overlapping augmented patches and sample $N = n \cdot k$ positive pairs of voxels from their overlapping regions, i.e., pairs of voxels from different patches but having the same absolute positions $\{p_i\}_{i = 1}^N$ in the original image; see Figure~\ref{fig:ape}. Patch augmentations include rescaling to random spacing, random masking, random gaussian blur/sharpening/noising, and clipping the intensities to random Hounsfield window. Then, we predict the APE embeddings $\{(a_i, a_i^+)\}_{i = 1}^N$ for all the $N$ positive pairs of voxels and compute the pairwise distances $d^{\text{pred}}_{ij} = \|a_i - a_j^+\|_2$. The target distances $d^{\text{true}}_{ij} = \|\hat{p}_i - \hat{p}_j\|_2$ are computed as before, between the normalized versions $\{\hat{p}_i\}_{i = 1}^N$ of the absolute positive pairs' coordinates $\{p_i\}_{i = 1}^N$. Note that the distances $d^{\text{pred}}_{ij}$ between voxels from the same overlapping region are local, but their predictions $d^{\text{pred}}_{ij} = \|a_i - a_j^+\|_2$ are now computed between the embeddings from the different patches, which prevents trivial solutions and improves the APE properties at local scale. The modified training objective is given by

\begin{equation}
    L = L_{\text{dist}} + \lambda \cdot L_{\text{equiv}} = \frac{1}{N^2}\sum_{i = 1}^N\sum_{j = 1}^N (d^{\text{pred}}_{ij} - d^{\text{true}}_{ij})^2 + \lambda \cdot \frac{1}{N} \sum_{i = 1}^N (d^{\text{pred}}_{ii})^2
\end{equation}

\noindent
The second term $L_{\text{equiv}}$ additionally enforces equivariance of the APE embeddings w.r.t. crops and patch augmentations. In our experiments, we ablate its effect by comparing model with $\lambda = 0.0$ (\textit{APE-Aug}) and $\lambda = 1.0$ (\textit{APE-equiv}). 

\subsection{Evaluating APE on anatomical landmark retrieval}
\label{subsec:retrieval}

Following \cite{sam}, we assess how well the APE embeddings correspond to anatomical positions by applying them to anatomical landmark retrieval.

To do so, we label a voxel corresponding to a certain anatomical landmark on a train image and compute its APE embedding, which we refer to as \textit{query}. To retrieve a voxel with the closest anatomical position on a test image, we compute its APE map and perform a nearest neighbor search over this map to find the voxel with the closest APE embedding to the query (by euclidian distance).

As a quality metric we compute the radial error (in mms), i.e., the physical distance between the retrieved voxel and the labeled voxel corresponding to the ground truth location of the anatomical landmark of interest in the test image. We report the mean radial error (MRE) and the standard deviation of radial error over all possible train-test image pairs from the dataset.

\subsection{Evaluating APE on few-shot organ localization}
\label{subsec:localization}

Following~\cite{rpr}, we also evaluate APE in a more practice-oriented application -- weakly supervised few-shot organ localization. In this task, we aim to build a bounding box predictor for a certain organ, based on some cheap labeling (e.g., bounding boxes or key points) of few train images.

Idea of applying APE to this task is to define several anatomical landmarks from which the organ's bounding box can be constructed, and reduce the bounding box prediction problem to these landmarks retrieval. For a fair comparison with~\cite{rpr}, we use the same choice of these box-defining landmarks --- the organ's six edge points (two points along each of image axes). At the training stage we label them on a train image and precompute their APE embeddings. At the inference stage, we use each of the precomputed embeddings as a query to retrieve the voxel with the closest anatomical position on a test image, as described in Section~\ref{subsec:retrieval}. Then, the prediction is obtained as the bounding box of the retrieved voxels. To make this procedure few-shot, we repeat it for all train images and average the bounding box predictions on a test image.

As a quality metric, we compute the intersection over union (IoU) between the predicted box and the ground truth organ's box. In addition, we evaluate the bounding box predictor when it is used for cropping a raw image to the organ of interest, as a preprocessing step reducing the computational burden of further image analysis steps. In this scenario, it is crucially important to achieve almost total recall in order not to crop some part of the organ. The most simple way to increase recall is to enlarge the predicted boxes by $\alpha$ times w.r.t to their centers. For each organ, we choose the $\alpha$ such that recall w.r.t. to the organ masks becomes not less than $0.99$ and compute the ratios between the raw image volumes and the volumes of images cropped to the enlarged boxes. We refer to this metric as volume ratio at 0.99 recall (VR@99). We report mean and standard deviation statistics of IoU and VR@99 across the test images and cross-validation with train folds of size $5$ ($5$-shot learning).


\section{Experiments \& results}
\label{sec:results}

\subsection{Datasets \& implementation details}

We use three publicly available CT datasets for APE training. We incorporate 2400 CTs from the AMOS~\cite{amos} and 2000 CTs from the unlabeled part of the FLARE2022~\cite{flare}, both covering the abdominal domain. Additionally, we include 4000 chest CTs from the NLST~\cite{nlst}. The resulting training set is entirely unlabeled. It covers both the abdomen and thorax domains in a roughly equal proportions, establishing a foundation for the model's robustness in both areas.

We train the APE model for 150k batches of $n = 8$ pairs of overlapping patches with $N = 8000$ positive pairs of voxels. The training takes 17 hours on a single NVIDIA RTX A4000-16GB GPU. We use AdamW optimizer with a constant learning rate $0.0003$, weight decay $10^{-6}$ and gradient clipping to $1.0$ norm. Pre-processing includes only cropping to dense foreground voxels (thresholded by $-500$HU). Voxel spacing is randomly augmented in the range from $1 \times 1 \times 1.5$ to $2 \times 2 \times 3$ mm$^3$. Patch size is augmented such that patch volume is equal to $96 \times 96 \times 64$ voxels and aspect ratios is not larger than $2$.

To evaluate APE and baselines, we use the labeled part of FLARE2022 \cite{flare}. It consists of 50 CT scans along with segmentation masks of 13 abdominal organs. 

\subsection{Anatomical landmark retrieval}

We evaluate our APE models on retrieval of two types of anatomical landmarks: organs' centers of mass, and organs' edge points. We denote the corresponding mean radial errors as MRE$_c$ and MRE$_{ep}$, and report them together with inference time in Table~\ref{tab:ret}.
Note that the \textit{APE-Aug} is better than \textit{APE-Naïve} and \textit{APE-Equiv} is better than \textit{APE-Aug}, which confirms our intuition described in Section~\ref{sec:method}. Compared to RPR~\cite{rpr}, \textit{APE-Equiv} achieves the same performance by using much simpler one-stage retrieval procedure, while RPR relies on Monte-Carlo ensembling and uses two-stage pipeline with separate coarse and fine-grained models. Surprisingly, SAM yields the inferior performance in our experiments, despite that the authors report very good retrieval results in the original paper~\cite{sam}. We used their pre-trained weights and code, so our best guess is that the performance have dropped due to a domain shift between the FLARE2022 and the SAM training data.

\begin{table}
\caption{Mean radial errors and inference time of anatomical landmark retrieval.}
\centering
\begin{tabular}{l@{\hspace{10pt}}c@{\hspace{10pt}}c@{\hspace{10pt}}c@{\hspace{10pt}}c@{\hspace{10pt}}c}
    
    \toprule
                    & SAM \cite{yan2022sam} &  RPR \cite{rpr} & APE-Naïve & APE-Augm & APE-Equiv \\
    MRE$_c$, mm     & $62\pm34$ & $22\pm 15$ & $47 \pm 38$ & $32\pm31$ & $22\pm19$ \\
    MRE$_{ep}$, mm  & $73\pm42$ & $33\pm 30$ & $53\pm43$ & $41\pm40$ & $34\pm35$ \\
    Time, s         & 0.50 & 0.15 & 0.07 & 0.07 & 0.07 \\

     \bottomrule

\end{tabular}

\label{tab:ret}
\end{table}


    
         




\subsection{Few-shot organ localization}

IoU and VR@99 of few-shot organ localization for our APE models and baselines are shown in Tables~\ref{tab:iou},~\ref{tab:volume-ratios}. We employed nnUNet~\cite{nnunet} and nnDetection~\cite{nndet} as standard supervised baselines. Note that we assume that only weak labels of organs are available in the training set. Therefore, we use the filled bounding boxes as masks when training nnUNet. Both nnUNet and nnDetection performed worse than APE, which we explain by their weak few-shot capabilities.

Models producing 1D anatomical positional embeddings~\cite{bpr,deepindex} show poor results because they are able to predict bounding boxes along the only one axis.

Our \textit{APE-Equiv} model outperforms all the other models, including SAM~\cite{sam} and RPR~\cite{rpr}, with p-value $< 10^{-6}$ of Wilcoxon signed-rank test.

\begin{table}
\caption{IoU of few-shot localization of 13 organs on FLARE2022~\cite{flare} and average IoU across organs.}

\centering
\resizebox{\textwidth}{!}{%
\begin{tabular}{l@{\hspace{10pt}}c@{\hspace{10pt}}c@{\hspace{10pt}}c@{\hspace{10pt}}c@{\hspace{10pt}}c@{\hspace{10pt}}c@{\hspace{10pt}}c@{\hspace{10pt}}c@{\hspace{10pt}}c@{\hspace{10pt}}c@{\hspace{10pt}}c@{\hspace{10pt}}c@{\hspace{10pt}}c@{\hspace{10pt}}c}
    \toprule
    & LIV & RKI & SPL & PAN & AOR & IVC & RAG & LAG & GBL & ESO & STO & DUO & LKI & \textit{avg} \\
    \midrule

    nnUNet~\cite{nnunet} & 0.49 & 0.39  & 0.43  & 0.39  & 0.46  & 0.47  & 0.45  & 0.46  & 0.40  & 0.50  & 0.42  & 0.43 & 0.40  & 0.44 \\

    nnDetection~\cite{nndet} & 0.20  & 0.48  & 0.57 & 0.45  & 0.21  & 0.27  & 0.50  & 0.54  & 0.47  & 0.52  & 0.33  & 0.49  & 0.49  & 0.43 \\

    \midrule

    SAM~\cite{sam} & 0.61  & 0.37  & 0.35 & 0.43  & 0.38  & 0.37  & 0.17 & 0.17  & 0.19 & 0.27  & 0.48  & 0.40 & 0.35  & 0.35 \\


    BPR~\cite{bpr} & 0.18 & 0.02 & 0.03  & 0.05 & 0.02 & 0.01 & 0.00  & 0.00 & 0.01 & 0.01 & 0.07 & 0.03  & 0.02  & 0.04 \\

    Deep-index~\cite{deepindex} & 0.18 & 0.02 & 0.03 & 0.05 & 0.02  & 0.01  & 0.00 & 0.00  & 0.01 & 0.01 & 0.07 & 0.03 & 0.02 & 0.04 \\

    RPR~\cite{rpr} & \textbf{0.68} & 0.49  & \textbf{0.51} & 0.53 & 0.56  & 0.51 & 0.31 & 0.34  & 0.25 & 0.43 & \textbf{0.56} & 0.53 & 0.52  & 0.48 \\

    \midrule

    APE-Naïve & 0.62 & 0.42 & 0.46 & 0.46 & 0.53 & 0.55 & 0.28 & 0.25  & 0.18 & 0.44 & 0.50 & 0.43 & 0.45 & 0.43 \\
    
    APE-Augm & 0.66 & 0.48 & 0.48  & 0.49 & \textbf{0.59} & 0.60 & 0.33  & 0.31 & 0.24 & \textbf{0.49} & 0.55 & 0.52 & 0.52 & 0.48 \\
    
    APE-Equiv & 0.67 & \textbf{0.51} & 0.50 & \textbf{0.58} & \textbf{0.59} & \textbf{0.61} & \textbf{0.37} & \textbf{0.40} & \textbf{0.28}  & 0.46  & \textbf{0.56} & \textbf{0.57} & \textbf{0.56} & \textbf{0.51} \\
    
    \bottomrule
\end{tabular}}

\label{tab:iou}
\end{table}

\begin{table}
\caption{VR@99 of organ localization on FLARE2022~\cite{flare}.}

\centering
\resizebox{\textwidth}{!}{%
\begin{tabular}{l@{\hspace{10pt}}l@{\hspace{15pt}}l@{\hspace{15pt}}l@{\hspace{15pt}}l@{\hspace{15pt}}l@{\hspace{15pt}}l@{\hspace{15pt}}l@{\hspace{15pt}}l@{\hspace{15pt}}l@{\hspace{15pt}}l@{\hspace{15pt}}l@{\hspace{15pt}}l@{\hspace{15pt}}l@{\hspace{15pt}}l}

    \toprule
    & LIV & RKI & SPL & PAN & AOR & IVC & RAG & LAG & GBL & ESO & STO & DUO & LKI & $avg$ \\
    \midrule

    nnUNet~\cite{nnunet} & 4.2 & 32.7 & 28.9 & 17.4 & 28.9 & 50.3  & 378.1 & 371.3 & 1.0 & 193 & 11.9  & 34.5 & 1.0 & 88.71  \\

    nnDetection~\cite{nndet} & 1.9  & 1.0  & 1.0 & 1.8  & 1.0 & 1.0  & 1.0  & 1.0  & 1.0  & 1.0  & 1.0 & 2.9 & 1.0 & 1.28  \\

    \midrule

    SAM~\cite{sam} & 3.3  & 10.5 & 7.6 & 8.3 & 6.5 & 12.7  & 1.0  & 1.0  & 1.0 & 1.0  & 4.6 & 9.8  & 10.2 & 5.96 \\


    BPR~\cite{bpr} & 1.3  & 1.6  & 1.7 & 2.1 & 1.1 & 1.1 & 2.7  & 3.2 & 1.5 & 2.8  & 1.6 & 1.9 & 1.7 & 1.87 \\
    
    Deep-index \cite{deepindex} & 1.3 & 1.6 & 1.7 & 2.0  & 1.1 & 1.1 & 2.6  & 3.1 & 1.8 & 3.0  & 1.7 & 1.8 & 1.7 & 1.88 \\

    RPR~\cite{rpr} & 3.7  & \textbf{16.2}  & 11.6 & 14.0  & 25.8 & 54.6 & 67.0 & 66.0 & 7.3  & 38.6 & 4.7  & 22.1 & 20.5 & 27.08 \\

    \midrule

    APE-Naïve & 3.4  & 11.6  & 8.3 & 10.9 & 24.7 & 51.1  & 85.2 & 76.0 & 1.0  & 63.5  & \textbf{8.5} & 7.8 & 16.0 & 28.31 \\
    
    APE-Augm & \textbf{4.3} & 13.6 & 10.6 & 6.1 & 31.7  & 56.6 & 94.3 & 38.7 & 1.0 & \textbf{147} & 7.3 & 18.6 & 17.8 & 34.43\\
    
    APE-Equiv & 4.2 & 9.3 & \textbf{13.2}  & \textbf{23.3} & \textbf{34.1} & \textbf{58.5} & \textbf{141.}7 & \textbf{204.5} & \textbf{17.1} & 94.7 & 7.7 & \textbf{31.3} & \textbf{32.3} & 51.68 \\
    
    \bottomrule
    
\end{tabular}}

\label{tab:volume-ratios}
\end{table}

\subsection{Qualitative results}

\begin{figure}
     \centering
     \begin{minipage}[l]{0.53\textwidth}
         \includegraphics[width=\textwidth]{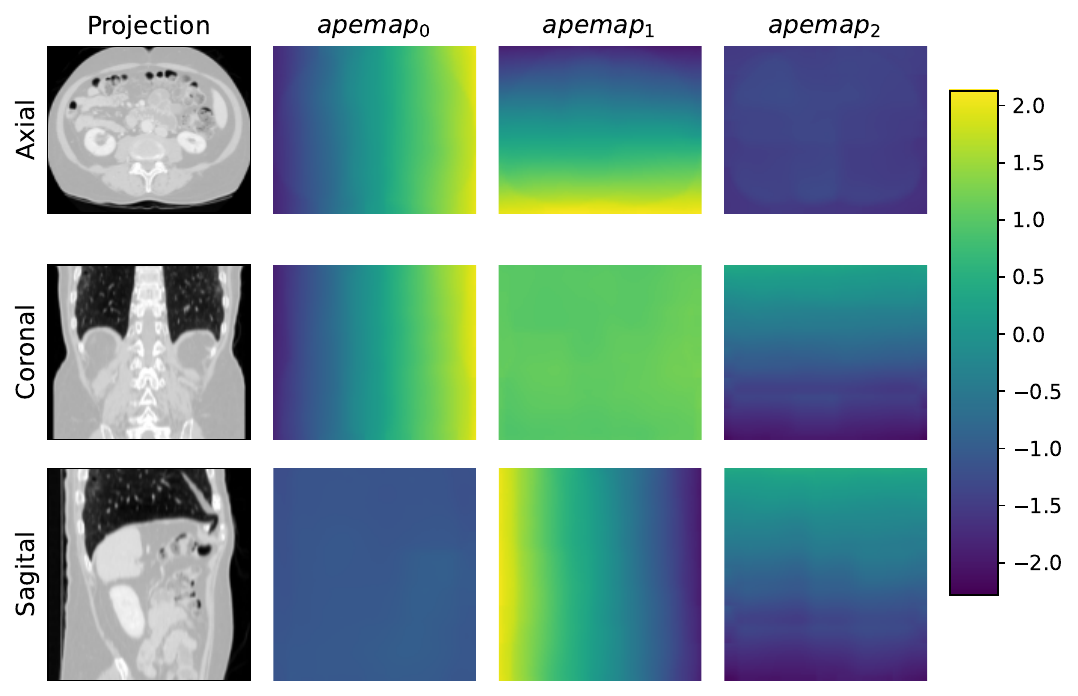}      
         \captionof{figure}{Continuous property of APE embedding maps}
         \label{fig:apecontinuous}
     \end{minipage}\hfill
     \begin{minipage}[r]{0.45\textwidth}
         \centering
         \includegraphics[width=\textwidth]{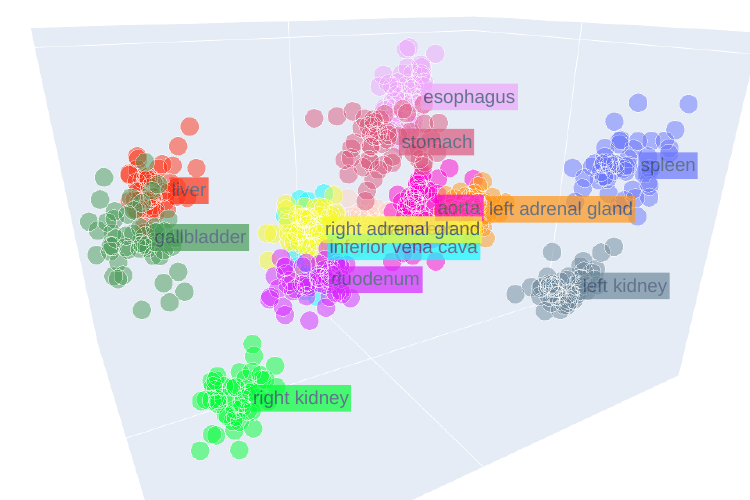}
         \captionof{figure}{Projection of 3D scatter plot of APE embeddings of different organs' centers}
         \label{fig:apeprojections}
     \end{minipage}
\end{figure}

Figure~\ref{fig:apecontinuous} shows an image along with the 3 channels of its APE map in axial, coronal and sagittal projections. Despite the patch-wise inference, APE embeddings are continuous w.r.t. voxels' physical coordinates. 

We visualize how different organs and anatomical landmarks are located in the 3D APE embeddings' space, by building a 3D scatter plot of the APE embeddings of the center voxels of 13 abdominal organs in 50 patients from FLARE2022~\cite{flare}; see Figure~\ref{fig:apeprojections}. Note that these embeddings turn out to be clusterized, and the relative location of clusters reminds the relative positions of abdominal organs.

\section{Conclusion}

In this work we propose a model of 3D anatomical positional embeddings. Our experiments showed that APE is SotA on anatomical landmark retrieval and few-shot organ localization tasks. The only existing model that competes with APE is RPR~\cite{rpr}. However, in contrast to RPR, APE is capable to produce embeddings for all the voxels of an input image, which simplifies the anatomical landmark retrieval pipeline and make APE more convenient and efficient for other applications.

There are several limitations of the current APE models. First, we have not managed to make them equivariant to flips and rotations. Second, they are trained only on abdominal and chest images and evaluated only on abdomen.

\begin{credits}

\subsubsection{\discintname}
The authors have no competing interests to declare that are
relevant to the content of this article.
\end{credits}

\bibliographystyle{splncs04}
\bibliography{bib}
\end{document}